\documentclass[letterpaper]{article} 
\usepackage{aaai25}  
\usepackage{times}  
\usepackage{helvet}  
\usepackage{courier}  
\usepackage[hyphens]{url}  
\usepackage{graphicx} 
\usepackage{natbib}  
\usepackage{caption} 
\usepackage{algorithm}
\usepackage{algorithmic}
\usepackage{amsmath}
\usepackage{booktabs}
\usepackage{xcolor}
\usepackage{multirow}
\usepackage{graphicx}
\usepackage{pifont}
\newcommand{\checkmark}{\ding{51}}

\usepackage{newfloat}
\usepackage{listings}
\DeclareCaptionStyle{ruled}{labelfont=normalfont,labelsep=colon,strut=off} 
\floatstyle{ruled}
\newfloat{listing}{tb}{lst}{}
\floatname{listing}{Listing}
\title{PointCFormer: a Relation-based Progressive Feature Extraction Network for \\
Point Cloud Completion}
\author{
    Yi Zhong\textsuperscript{\rm 1},
    Weize Quan\textsuperscript{\rm 2,3},
    Dong-Ming Yan\textsuperscript{\rm 2,3},
    Jie Jiang\textsuperscript{\rm 1}\thanks{Corresponding author: Jie Jiang},
    Yingmei Wei\textsuperscript{\rm 1}
}
\affiliations{
    \textsuperscript{\rm 1}National University of Defense Technology\\
    \textsuperscript{\rm 2}MAIS, Institute of Automation, Chinese Academy of Sciences
    \textsuperscript{\rm 3}University of Chinese Academy of Sciences\\

    \{zhongyi, jiejiang, weiyingmei\}@nudt.edu.cn,
    \{qweizework, yandongming\}@gmail.com
}

\usepackage{bibentry}

\begin{document}

\maketitle

\begin{abstract}
Point cloud completion aims to reconstruct the complete 3D shape from incomplete point clouds, and it is crucial for tasks such as 3D object detection and segmentation. Despite the continuous advances in point cloud analysis techniques, feature extraction methods are still confronted with apparent limitations. The sparse sampling of point clouds, used as inputs in most methods, often results in a certain loss of global structure information. Meanwhile, traditional local feature extraction methods usually struggle to capture the intricate geometric details. To overcome these drawbacks, we introduce PointCFormer, a transformer framework optimized for robust global retention and precise local detail capture in point cloud completion. This framework embraces several key advantages. First, we propose a relation-based local feature extraction method to perceive local delicate geometry characteristics. This approach establishes a fine-grained relationship metric between the target point and its k-nearest neighbors, quantifying each neighboring point's contribution to the target point's local features. Secondly, we introduce a progressive feature extractor that integrates our local feature perception method with self-attention. Starting with a denser sampling of points as input, it iteratively queries long-distance global dependencies and local neighborhood relationships. This extractor maintains enhanced global structure and refined local details, without generating substantial computational overhead. Additionally, we develop a correction module after generating point proxies in the latent space to reintroduce denser information from the input points, enhancing the representation capability of the point proxies. PointCFormer demonstrates state-of-the-art performance on several widely used benchmarks. Our code is available at https://github.com/Zyyyyy0926/PointCFormer\_Plus\_Pytorch.
\end{abstract}

%

\section{Introduction}

Point cloud is a common data format in 3D vision tasks, and point cloud-based analysis tasks have greatly promoted the development of 3D computer vision~\cite{mao2024denoising}. Point cloud data is generally obtained through 3D sensors~\cite{bi2010advances}, however, due to inevitable external factors such as occlusion, light reflection, and limited sensor resolution~\cite{chen2016dynamic}, as well as internal factors such as the inherent shape deficiencies of the object itself, point cloud data often exhibits incomplete and sparse characteristics. Point cloud completion is restoring this incomplete and sparse 3D point cloud data to a point cloud representing the complete shape of the object.

\begin{figure}[t]
\centering
\includegraphics[width=\columnwidth]{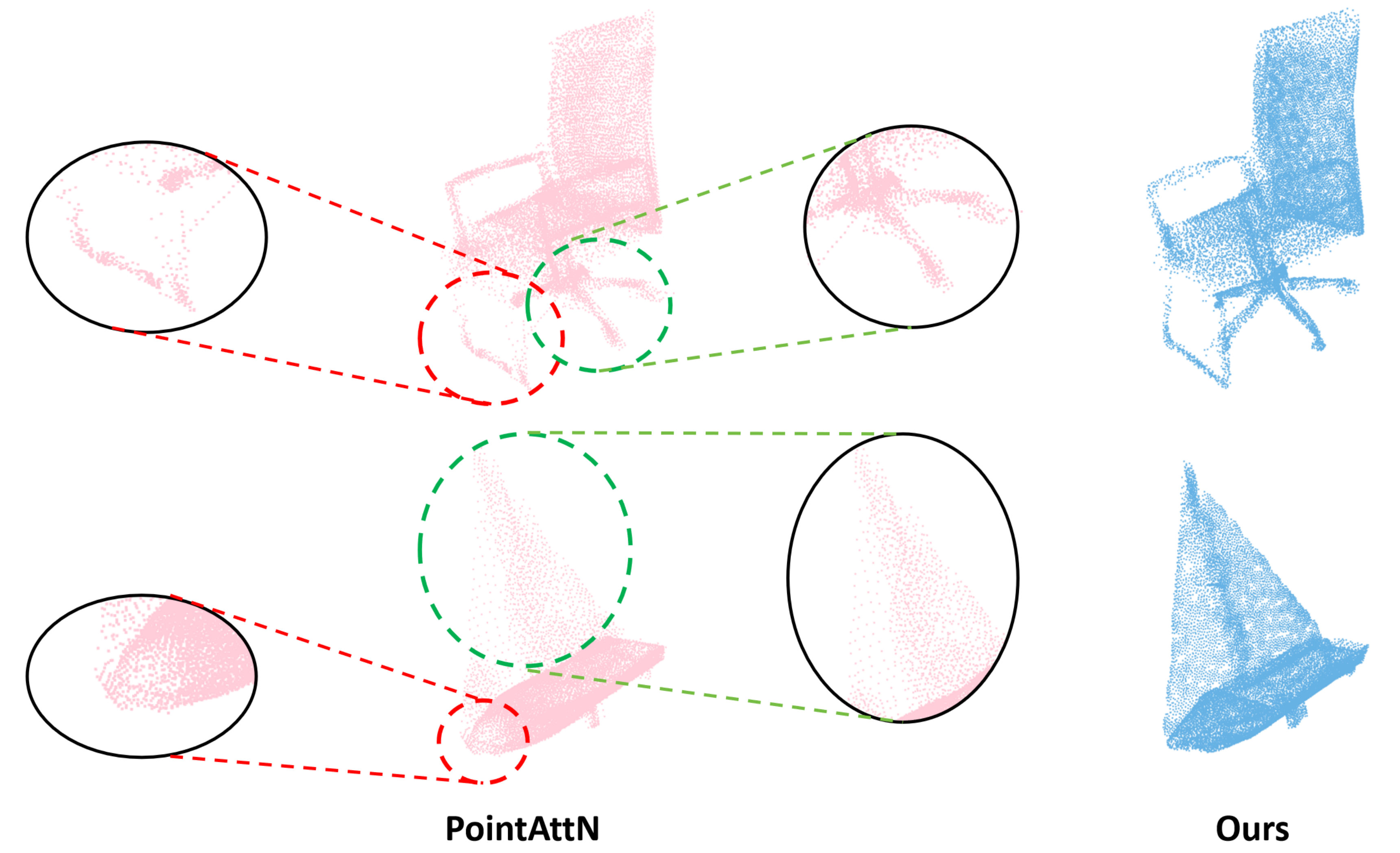}
\caption{Improved global shape and enriched local detail in point cloud completion using our method.}
\label{fig_show1}
\end{figure}

Recent deep learning advances have solidified the encoder-decoder framework as a staple for point cloud completion, with PointNet~\cite{qi2017pointnet} and PointNet++~\cite{qi2017pointnet++} laying the foundation for the feature encoder. The pioneering PointTr~\cite{yu2021pointr} leverages the Transformer model to enhance this architecture. SeedFormer~\cite{zhou2022seedformer} introduces an innovative shape representation for feature integration, while ~\cite{wang2022learning} focuses on local feature grouping to improve completion. FBNet~\cite{yan2022fbnet} combines feedback loops and cross transformers to better link feature levels, while SnowflakeNet~\cite{xiang2021snowflakenet} employs skip transformers to inject spatial relationships into the decoding stage. AnchorFormer~\cite{chen2023anchorformer} introduces anchors for region differentiation and predicts the complete shape by merging these anchors with the observed input points. AdaPoinTr~\cite{10232862} improves the performance based on PointTr, while the Mamba structure in 3DMambaComplete~\cite{li20243dmambacomplete} shows advantages in this field and pushes the envelope in point cloud completion.

Among these existing methods, the feature extraction of points is an essential step, which primarily relies on the combined use of down-sampling and k-nearest neighbor (k-NN) search. However, the risk of losing certain global structural information is inevitable if an overly sparse down-sampled point cloud is employed as the model input. In addition, traditional feature extraction algorithms based on simple k-NN search can result in the network only extracting relatively coarse local features. Unfortunately, these fundamental limitations have seldom been discussed in previous works.

In this paper, we delve into the utilization of an advanced feature extractor for point cloud completion, aiming to mitigate the loss of global information during point cloud sampling and the coarse extraction issues in local feature extraction. We propose a framework, termed PointCFormer, as shown in Fig.~\ref{fig_arc}. Specifically, PointCFormer adopts the transformer-based encoder-decoder architecture. To handle the coarse extraction problem, we propose fine-grained computation for the contribution of the neighboring point to the target point according to the geometry characteristics of the local region (including 3D spatial coordinates and high-dimensional feature vectors). To better preserve the global shape information during the sampling process, we introduce a progressive method with slow down-sampling, which gradually refines the features through alternately global and local queries. Compared to previous work, Fig.~\ref{fig_show1} shows the superior performance of our method in the point cloud completion task in terms of global shape and local details. In summary, our main contributions are as follows:
\begin{itemize}
\item We introduce a \emph{local geometric relationship perception} module, which adaptively determines point contribution weights within neighborhoods based on relation metrics in 3D space and high-dimensional feature space, thereby enhancing local feature extraction accuracy.
\item We propose a \emph{progressive feature extractor} that synergistically combines the advantages of kNN-based local perception and self-attention mechanism. The features are progressively refined through repeated alternation between global and local queries.
\item We devise a \emph{point proxy correction} module in the latent space to enhance the affinity between the generated point proxies and the original input point cloud, thereby ensuring that the generated proxies are more closely aligned with the input data.
\item Our PointCFormer achieves state-of-the-art performance on four common datasets for point cloud completion.
\end{itemize}

\section{Related Work}
\subsection{3D Point Cloud Completion} For 3D point cloud completion, traditional approaches~\cite{dai2017shape,han2017high,stutz2018learning} employ voxelization or distance fields to describe 3D objects and process them using 3D convolutional neural networks~\cite{wu20153d}. However, the high demand for memory and computational resources limits their application scope~\cite{wang2022learning}. To mitigate this issue, researchers have shifted towards using unstructured point clouds to represent 3D objects, exploring various innovative approaches. PointNet and its variants~\cite{qi2017pointnet,qi2017pointnet++} extract features directly from unstructured point clouds, provide a new perspective on the task of 3D point cloud processing. PCN~\cite{yuan2018pcn} leverages an encoder-decoder architecture and simulates the deformation process of the 2D plane through the FoldingNet technique~\cite{yang2018foldingnet}. This maps 2D points onto 3D surfaces to achieve point cloud completion while preserving geometric structure and topological properties. SnowflakeNet~\cite{xiang2021snowflakenet} models the point cloud generation process as a snowflake-like growth pattern based on certain base points in 3D space. This pattern gradually expands from the parent point to form a complete 3D shape, simulating complex geometric structures. LAKeNet~\cite{tang2022lake} introduces a novel 3D shape prediction method, which effectively captures the key features of 3D shapes by considering the structured and topological information of the predicted 3D shapes and following a key point-skeleton-shape prediction process. PointAttN~\cite{Wang_Cui_Guo_Li_Liu_Shen_2024} transforms the traditional k-NN method of obtaining local features into an attention-based method. This refines the global dependency on the relationships between point clouds. SeedFormer~\cite{zhou2022seedformer} introduces patch seeds as a new shape representation for point clouds and designs an upsampling transformer to make point cloud completion more efficient and accurate. Anchorformer~\cite{chen2023anchorformer} enhances the accuracy of point cloud completion by capturing region information through pattern recognition nodes and combining the information of sampling points and anchor points. Recently, \cite{li20243dmambacomplete} applies the Mamba~\cite{mamba} structure to point cloud completion, showcasing its advantages in processing 3D point cloud data, which brings new insights to the community.

\begin{figure*}[t]
\centering
\includegraphics[width=1\textwidth]{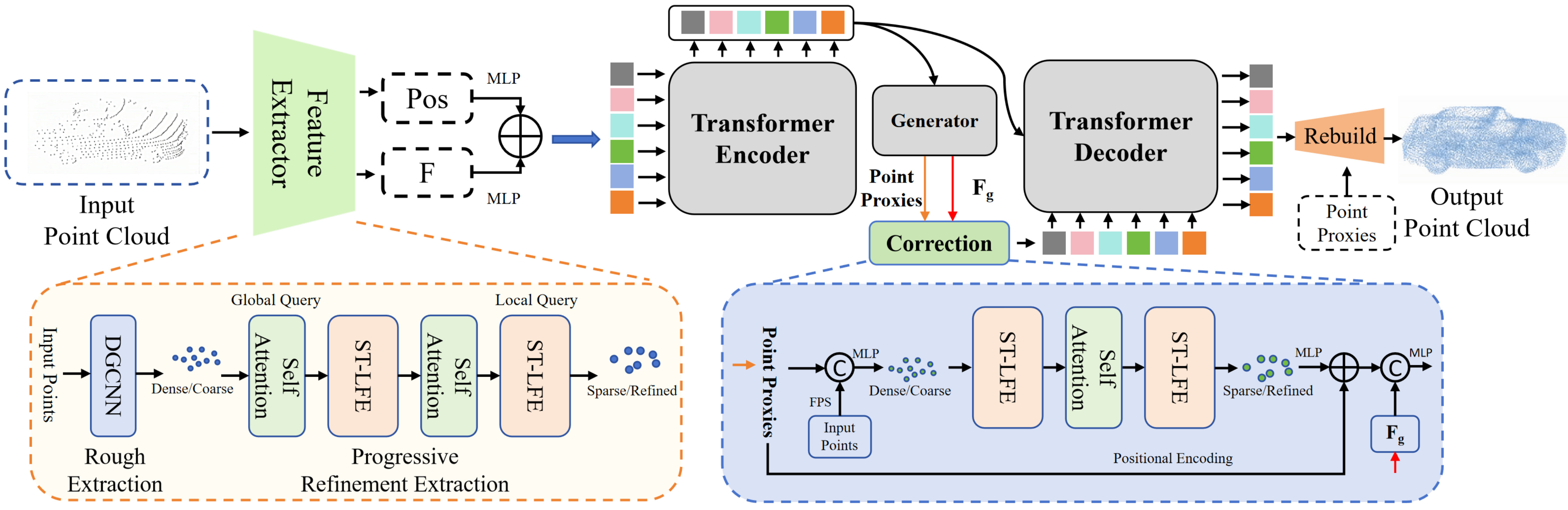}
\caption{Overview of PointCFormer framework: Initially, we extract representative sampling points and their local features from the input incomplete point cloud using a feature extractor. After adding position embeddings to the local features, we employ a Transformer encoder-decoder architecture to predict point proxies for the missing parts. Concurrently, a correction module aligns these point proxies with the original point cloud distribution. Finally, a simple MLP and a Rebuild head are used to complete the point cloud based on the predicted point proxies in a coarse-to-fine manner.}
\label{fig_arc}
\end{figure*}

\subsection{Transformer} In the field of natural language processing, the Transformer architecture~\cite{vaswani2017attention} has revolutionized sequence tasks with its self-attention and cross-attention mechanisms. In computer vision, Vision Transformer~\cite{dosovitskiy2020image,han2022survey} applies the Transformer architecture to 2D image processing, successfully managing image classification by deconstructing images into token sequences. Further research, like DeiT~\cite{touvron2021training}, has expanded the Transformer's efficient training strategy in visual tasks. However, applying the Transformer to 3D point cloud data is still exploratory~\cite{guo2021pct,wang2023octformer,misra2021end}. The unstructured, high-dimensional nature of such data presents certain challenges. Preliminary studies show that the Transformer can effectively capture local and global features in point clouds, providing a new approach to 3D point cloud recognition~\cite{mao2021voxel,pan20213d} and completion~\cite{wen2022pmp,zhang2022point,fei2023dctr}. This involves using the Transformer's self-attention mechanism to handle point cloud data's local structure and the cross-attention mechanism to predict missing parts using existing point cloud information.

\section{Proposed Method}
Fig.~\ref{fig_arc} outlines the PointCFormer architecture for point cloud completion. The process begins by feeding an incomplete point cloud into a feature extractor, where a single-layer DGCNN performs initial dense sampling and extracts coarse features. These features are then subjected to a progressive refinement process. Employing a self-attention module to model global correlations and the scale-tailored local feature extractor (ST-LFE) to capture local detail features, the features transform from a dense and coarse state to a sparse and refined one. These features are then spatially position-encoded and fed into an encoder. In the latent space, the generator\footnote{We follow the design of generator from \cite{10232862}.} uses the encoder's output features to create point proxies for the predicted 3D shape. Meanwhile, a correction module fuses these point proxies with spatial information from the dense input, ensuring alignment with the original point cloud distribution. Finally, the decoder outputs are used to gradually complete the point cloud with the help of an MLP and a reconstruction head.

\begin{figure}[t]
\centering
\includegraphics[width=1\columnwidth]{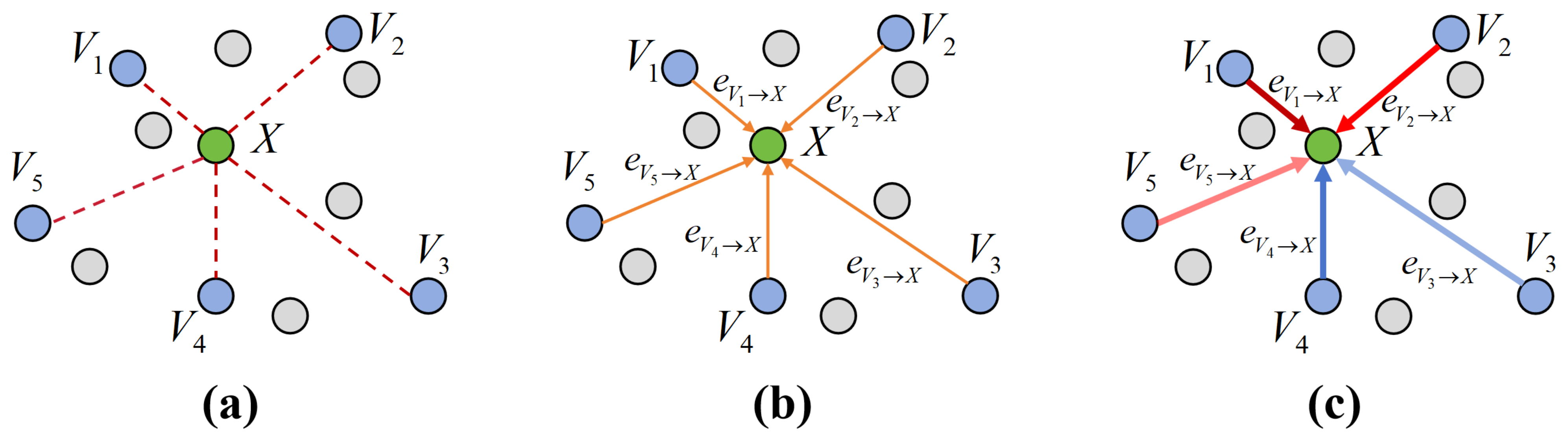}
\caption{Comparison of local feature extraction: traditional kNN-based method (Left); DGCNN with relative relation fusion (Center); Ours (Right). In (c), the redder the line segment, the stronger the correlation between the two connected points; the bluer the line segment, the weaker the correlation.}
\label{fig_need}
\end{figure}

\subsection{Local Geometric Relationship Perception}

\begin{figure}[t]
\centering
\includegraphics[width=\columnwidth]{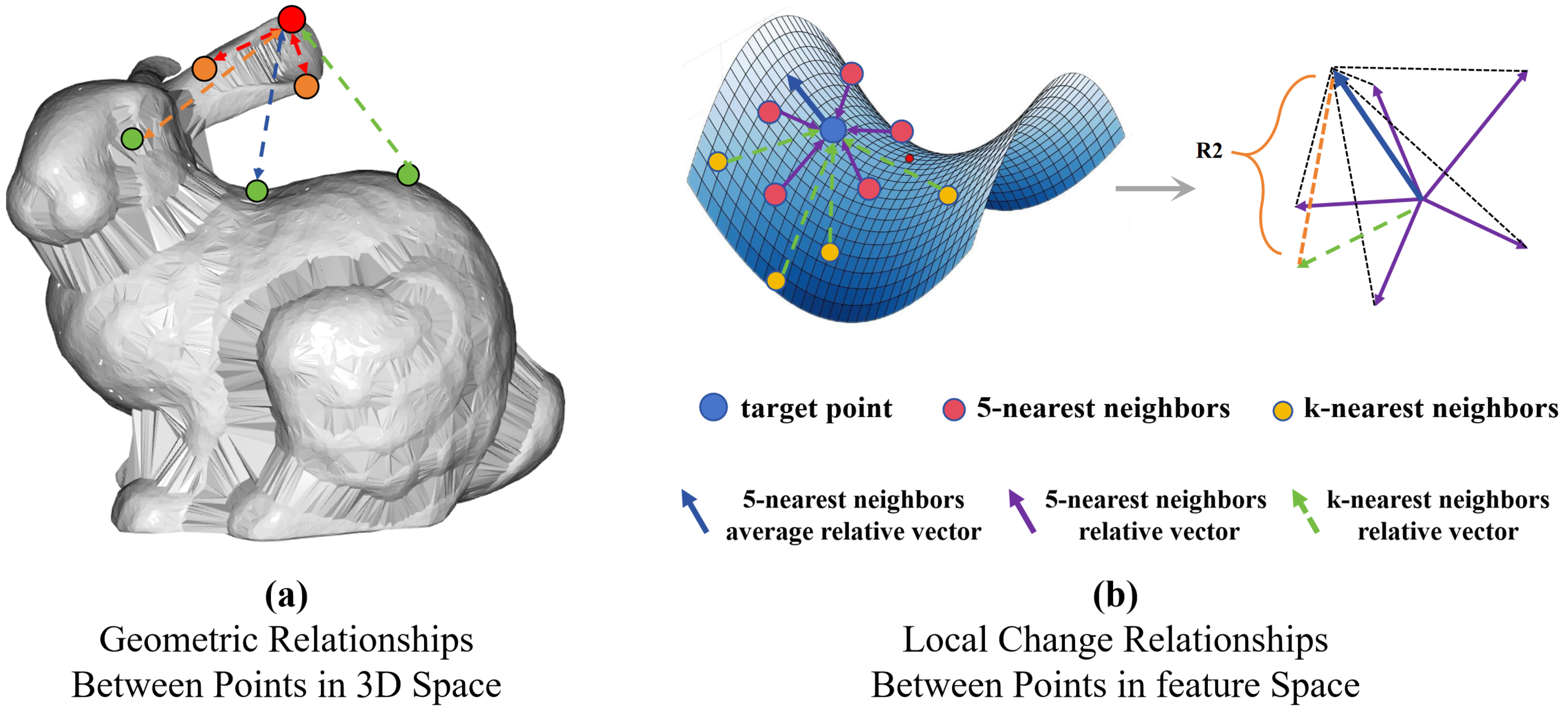}
\caption{Quantification of the relationship between two points.}
\label{fig_r1r2}
\end{figure}

Regular convolution operators~\cite{lecun1998gradient} on images cannot be directly applied to point cloud data. To address this challenge, many existing point cloud completion methods typically adopt a kNN-based method (Fig.~\ref{fig_need}(a)), which is essentially a variant of PointNet. This approach identifies local neighborhood points via k-NN search, maps these points' features to a high-dimensional space using an MLP, and then employs maximum pooling to extract local features. During the aggregation process, however, the relative positional relationships between points may be overlooked, thus limiting its effectiveness in capturing detailed local structural information. The DGCNN method~\cite{wang2019dynamic} (Fig.~\ref{fig_need}(b)) improved this only by incorporating the difference between feature vectors of points (directed edges in Fig.~\ref{fig_need}(b)) into the MLP learning process. However, these two methods still have limited capability for local geometric relationship perception and cannot fully mitigate the impact of the point cloud's sparsity and incompleteness.

\begin{figure}[t]
\centering
\includegraphics[width=0.95\columnwidth]{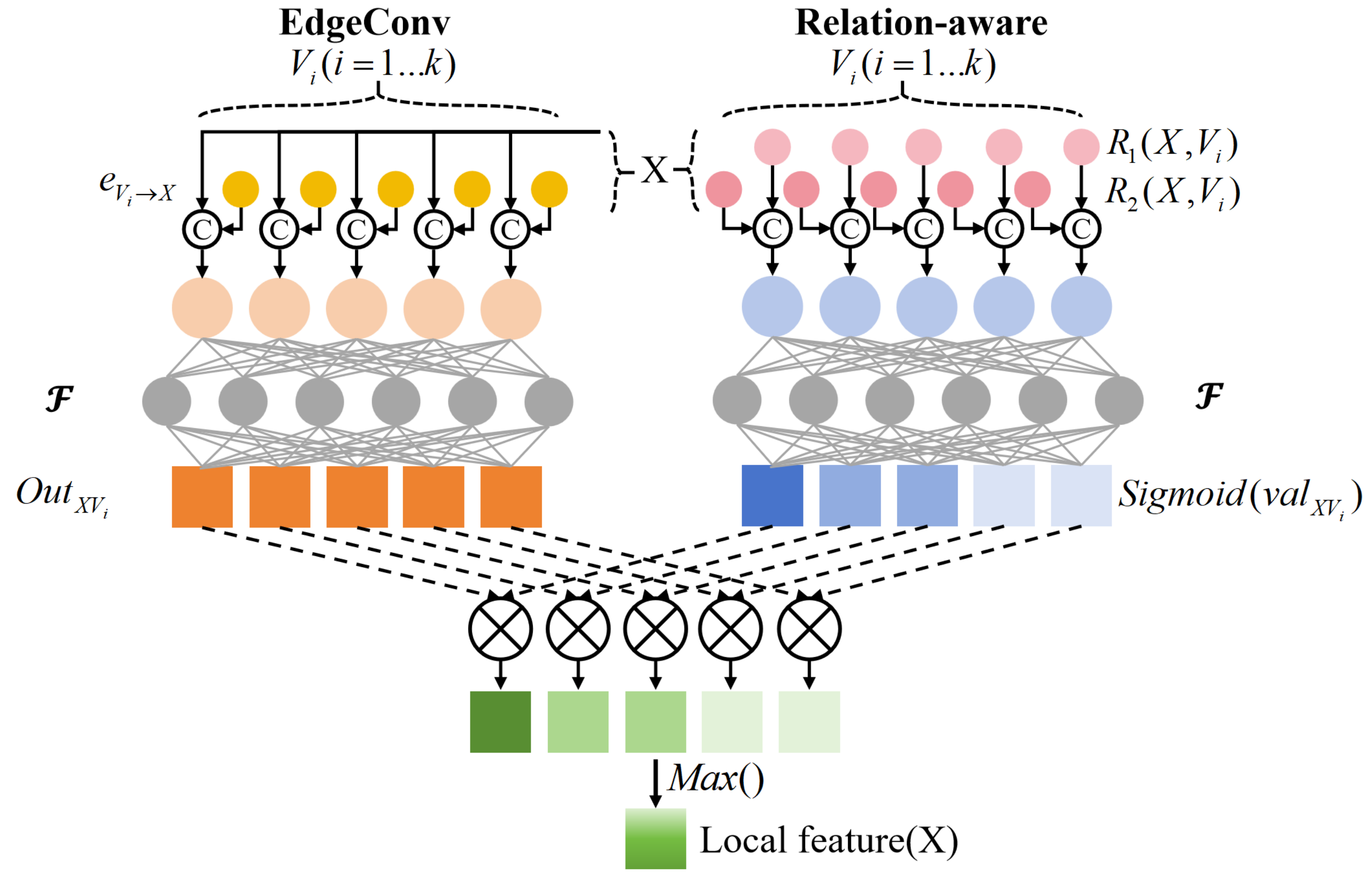}
\caption{The framework of local geometric relationship perception.}
\label{fig_local}
\end{figure}

The k-NN search may identify neighbors with inaccurate relevance to the target points due to sparse sampling, as illustrated in Fig.~\ref{fig_r1r2}(a), where only the two closest points (orange) show significant correlation with the target points (red), and more distant points (green) may provide little or even misleading information. To solve this problem, we propose a scheme (Fig.~\ref{fig_need}(c)) that introduces an adaptative contribution weight for each nearest neighbor based on its correlation with the target point.  Specifically, we quantify the interrelationships between a target point and its neighboring points in terms of geometry in 3D space and feature vectors in high-dimensional space simultaneously. The quantified values are then utilized in a learning-based approach to assess each neighbor's contribution to the target point. 

As previously discussed and substantiated by the examples presented in Fig.~\ref{fig_r1r2}(a), the spatial distance between two points significantly impacts their geometric correlation. Accordingly, we use the Manhattan distance between the target point \(Q\) and its nearest neighbor point \(V_i\) (\(i = 1, \ldots, k\)) in different directions as the spatial relationship metric \(R1\). This is formulated as follows:
\begin{equation}
R_1(Q, V_i)=\left(\left|x_Q-x_{V_i}\right|,\left|y_Q-y_{V_i}\right|,\left|z_Q-z_{V_i}\right|\right).
\end{equation}

To measure the relationship between points in high-dimensional feature space, a simple method is to directly extend the \(R1\) metric to a high dimension. Nevertheless, it does not produce satisfactory results. Therefore, we design a more intricate yet effective relationship metric as shown in Fig~\ref{fig_r1r2}(b), denoted as \(R2\). Initially, we select a small subset (M points) of the nearest neighbors and calculate their average relative relationship (average of directed edges) with the target point. This average vector represents the \emph{primary} change trend of the target point within a small region. We then subtract this average vector from the directed edge between the target point and each of its neighbors and take the absolute value to serve as the relationship metric between the two points. It's noteworthy that the \(R1\) is derived from the direct relation between points, while \(R2\) is derived from the relative relation (the edge between two points). The formulation of \(R2\) is as follows:
\begin{equation}
R_2\left(Q, V_i\right)=\left|\frac{1}{M} \sum_{j=1}^M \operatorname{vec}\left(Q, V_j\right)-\operatorname{vec}\left(Q, V_i\right)\right|,
\end{equation}
where $vec$ represents the vector of differences (directed edges) between two points.

Fig.~\ref{fig_local} illustrates the technical implementation of local geometric relationship perception. Specifically, the left part utilizes the process of EdgeConv for local feature extraction. The right part is our proposed relationship perception. We concatenate the relationship metrics \(R1\) and \(R2\) and use an MLP to output contribution weights in a learning manner. Then, the generated weights are element-wise multiplied by the extracted features. Finally, robust and geometry-related features are generated through a max function. 
We provide a more detailed analysis of local geometric perception with respect to the relationship metrics \(R1\) and \(R2\) in the Supplementary Material\cite{zhong2024pointcformer}.

\begin{figure}[t]
\centering
\includegraphics[width=0.85\columnwidth]{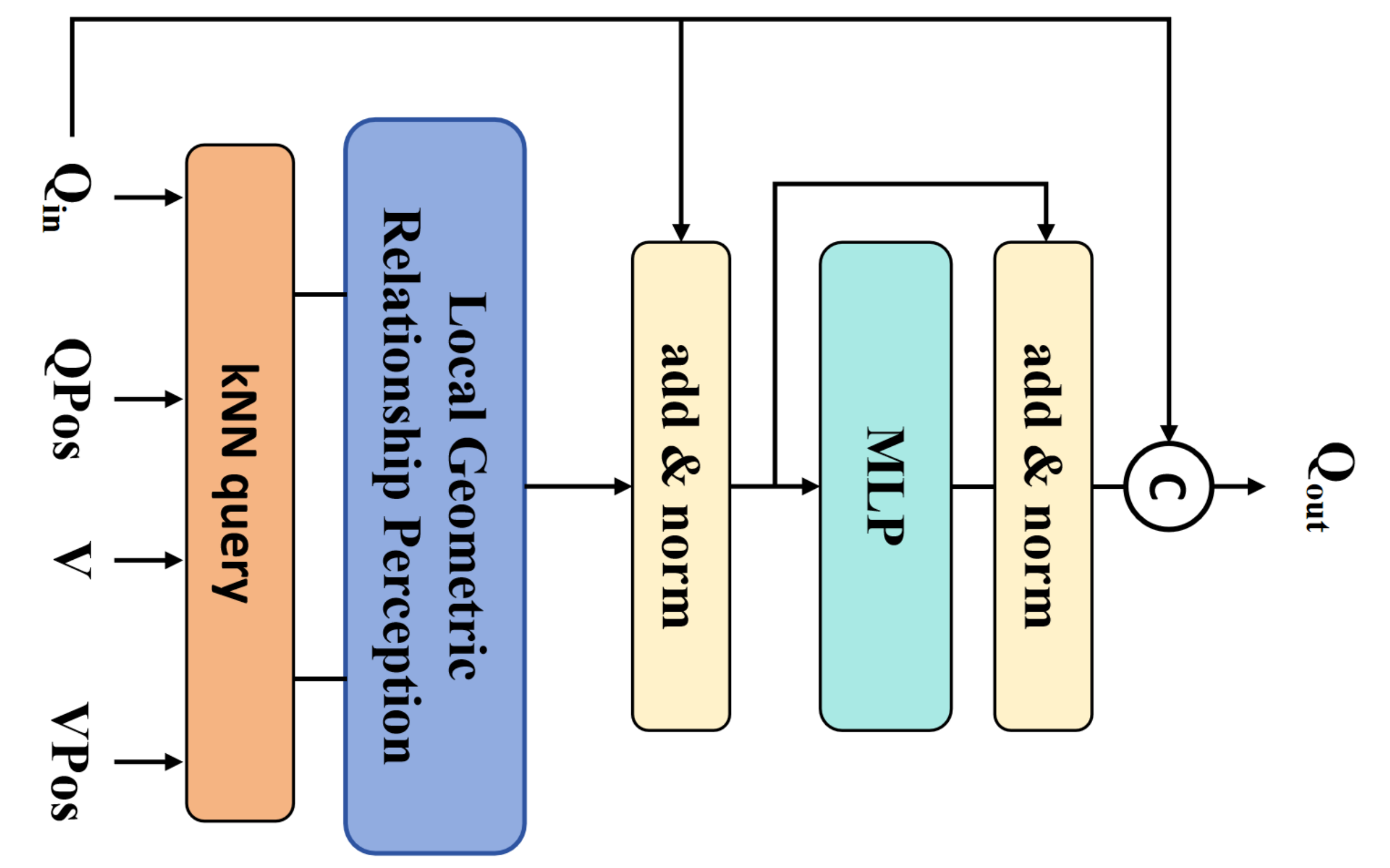}
\caption{Scale-tailored local feature extractor.}
\label{fig_atknn}
\end{figure}

\subsection{Progressive Feature Extractor}
\label{subsec:progressive}
Many point cloud completion approaches utilize the multi-layer DGCNN~\cite{yu2021pointr} consisting of EdgeConv and farthest point sampling (FPS) at each layer. This configuration allows for simultaneous down-sampling and feature extraction of the input incomplete point cloud. However, this approach fails to capture global point relationships. Recently, PointAttn~\cite{Wang_Cui_Guo_Li_Liu_Shen_2024} introduced a purely attention-based feature extraction method that establishes long-distance dependencies using a global self-attention mechanism, which is computationally expensive. To leverage the strengths of both approaches and incorporate our relation-based local geometric feature extraction, we develop a progressive feature extraction process that moves from coarse to fine granularity in the feature level while slowly reducing the point's number (from dense to sparse configurations), efficiently balancing computational cost and extraction accuracy.

As depicted in the lower left corner of Fig.~\ref{fig_arc}, our method is structured into two phases. In the initial phase, the input incomplete point cloud is subjected to a single-layer DGCNN for preliminary sampling and feature extraction, where we conducted a relatively dense sampling. Due to the single-layer operation, the features obtained at this stage are relatively coarse. The second phase involves a progressive refinement of feature extraction, with two alternating rounds of global relationship querying via self-attention modules and local relationship querying via the ST-LFE module. The design of ST-LFE enables both feature dimensionality expansion and point scale pruning, with its internal structure detailed in Fig.~\ref{fig_atknn}. Consequently, the dense and coarse features in the first phase are refined in the second phase, transforming into sparser but more intricately expressed sampling point features.

\subsection{Point Proxy Correction Module}
In previous studies, the feature of point proxies within the Transformer's decoder was facilitated through a cross-attention mechanism with the hidden features output by the encoder, enabling an implicit information exchange. While the point proxies predicted by the encoder are with high-dimensional features, their inherent sparsity falls short of capturing the dense distribution characteristics of the original 3D point cloud, thereby leading to a partial information loss. To enrich and enhance the representation of point agent features, we propose a correction module as shown in Fig.~\ref{fig_arc}. This module reuses a dense input point cloud, employing a processing flow akin to the feature extractor we have developed in Section~\ref{subsec:progressive}. Specifically, within a low-dimensional space, this module explicitly conveys dense point cloud information to the generated point proxies through self-attention and a local feature extractor. Consequently, point proxies are endowed with comprehensive information about the dense point cloud, more accurately reflecting the distribution characteristics of the original input data.

\subsection{Network Optimization}
For the loss function, we adopt a setting similar to AdaPoinTr. The first training objective is to minimize the Chamfer distance between the predicted point cloud (consisting of a sparse point proxy set called ``$\mathcal{C}$'' and a final dense prediction point set called ``$\mathcal{P}$'') and the actual point cloud (called ``$\mathcal{G}$''), thereby ensuring a closer approximation to the true point cloud configuration. It is formulated as:

\begin{equation}
J_0=\frac{1}{n_{\mathcal{C}}} \sum_{c \in \mathcal{C}} \min _{g \in \mathcal{G}}\|c-g\|+\frac{1}{n_{\mathcal{G}}} \sum_{g \in \mathcal{G}} \min _{c \in \mathcal{C}}\|g-c\| \text {, }
\end{equation}

\begin{equation}
J_1=\frac{1}{n_{\mathcal{P}}} \sum_{p \in \mathcal{P}} \min _{g \in \mathcal{G}}\|p-g\|+\frac{1}{n_{\mathcal{G}}} \sum_{g \in \mathcal{G}} \min _{p \in \mathcal{P}}\|g-p\| .
\end{equation}

The second training objective is a denoising task designed to enhance the model's robustness to random noise. Given a noisy query \(\hat{\mathcal{Q}}_i\) and corresponding noisy center \(\hat{c}_i^{gt} = n_i + c_i^{gt}\), the model aims to reconstruct the detailed local shape centered at \(c_i^{gt}\), despite the presence of noise \(n_i\). We denote the true local shape centered at \(c_i^{gt}\) as \(\mathcal{G}_{c_i}^{gt}\), and the local shape predicted from \(\hat{\mathcal{Q}}_i\) as \(\hat{\mathcal{P}}_i\), the auxiliary loss function can be expressed as follows:

\begin{equation}
J_{\text {denoise }}=\frac{1}{\left|\hat{\mathcal{P}}_i\right|} \sum_{c \in \hat{\mathcal{P}}_i} \min _{g \in \mathcal{G}_{c_i}^{gt}}\|c-g\|+\frac{1}{\left|\mathcal{G}_{c_i}^{g t}\right|} \sum_{g \in \mathcal{G}_{c_i}^{gt}} \min _{c \in \mathcal{P}_i}\|g-c\| \text {. }
\end{equation}

To this end, the final objective for our network is $J_{P C}=J_0+J_1+\lambda J_{\text {denoise }}$.

\section{Experiments}
\subsection{Datasets and Implementation Details}
We train PointCFormer on several datasets, including PCN~\cite{yuan2018pcn}, ShapeNet-55~\cite{yu2021pointr}, ShapeNet-34/Unseen-21~\cite{yu2021pointr} and KITTI. To ensure a fair comparison, we adhere to the standard protocols for training and testing on each dataset. PointCFormer is implemented with PyTorch and trained on two NVIDIA 4090 GPUs. The internal transformer encoder consists of six cascaded standard self-attention blocks, and the point cloud generation and transformer decoder scheme are similar to the Adapointr framework. Our network is trained using the AdamW optimizer with a base learning rate set to 0.0001. We use L1/L2 Chamfer distance and F-Score~\cite{10232862} as evaluation metrics for the PCN, ShapeNet-55, and ShapeNet34/Unseen21. On KITTI, we report the fidelity distance (FD) and minimum matching distance (MMD).

\subsection{Comparisons with State-of-the-Art Methods}
We compare our method with several advanced methods, including FoldingNet~\cite{yang2018foldingnet}, PCN~\cite{yuan2018pcn}, GRNet~\cite{xie2020grnet}, PoinTr~\cite{yu2021pointr}, SnowFlakeNet~\cite{xiang2021snowflakenet}, SeedFormer~\cite{zhou2022seedformer}, AnchorFormer~\cite{chen2023anchorformer}, PointAttN~\cite{Wang_Cui_Guo_Li_Liu_Shen_2024}, and AdaPointr~\cite{10232862}.

\begin{table*}[t]
\caption{Performance comparison on the PCN dataset. We use the CD-$\ell_1$(multiplied by 1000) and F-Score@1\% to compare with other methods.}
\centering
\resizebox{0.97\linewidth}{!}{
    \begin{tabular}{c|cccccccc|cc}
        \toprule[0.5mm]
        Methods & Plane & Cabinet & Car & Chair & Lamp & Sofa & Table & Boat & Avg CD-$\ell_1$ & F-Score@1\% \\
        \midrule
        FoldingNet(cvpr2019) & 9.49 & 15.80 & 12.61 & 15.55 & 16.41 & 15.97 & 13.65 & 14.99 & 14.31 & 0.322 \\
        PCN(3dv2018) & 5.50 & 22.70 & 10.63 & 8.70 & 11.00 & 11.34 & 11.68 & 8.59 & 9.64 & 0.695 \\
        GRNet(eccv2020) & 6.45 & 10.37 & 9.45 & 9.41 & 7.96 & 10.51 & 8.44 & 8.04 & 8.83 & 0.708 \\
        PoinTr(iccv2021) & 4.75 & 10.47 & 8.68 & 9.39 & 7.75 & 10.93 & 7.78 & 7.29 & 8.38 & 0.745 \\

        Snowflake(2021iccv) & 4.29 & 9.16 & 8.08 & 7.89 & 6.07 & 9.23 & 6.55 & 6.40 & 7.21 & - \\
        SeedFormer(eccv2022) & 3.85 & 9.05 & 8.06 & 7.06 & 5.21 & 8.85 & 6.05 & 5.85 & 6.74 & - \\
        AnchorFormer(cvpr2023) & 3.97 & 9.59 & 8.53 & 8.46 & 6.39 & 9.13 & 6.73 & 6.16 & 7.37 & - \\
        PointAttN(aaai2024) & 3.88 & 17.923 & 9.01 & 7.28 & 5.97 & - & - & - & 6.84 & - \\
        3DMambCom(arxiv2024) & 3.86 & 9.11 & 7.72 & 7.41 & 5.73 & 9.04 & 6.29 & 6.09 & 6.91 & 0.824 \\
        AdaPoinTr(tpami2023) & 3.68 & 8.82 & 7.47 & 6.85 & 5.47 & 8.35 & \textbf{5.80} & 5.76 & 6.53 & 0.845 \\
        \midrule
        PointCFormer & \textbf{3.53} & \textbf{8.73} & \textbf{7.32} & \textbf{6.68} & \textbf{5.12} & \textbf{8.34} & 5.86 & \textbf{5.74} & \textbf{6.41} & \textbf{0.855}\\
        \toprule[0.5mm]
\end{tabular}}
\label{table:pcn}
\end{table*}

\begin{figure*}[t]
\centering
\includegraphics[width=1\textwidth]{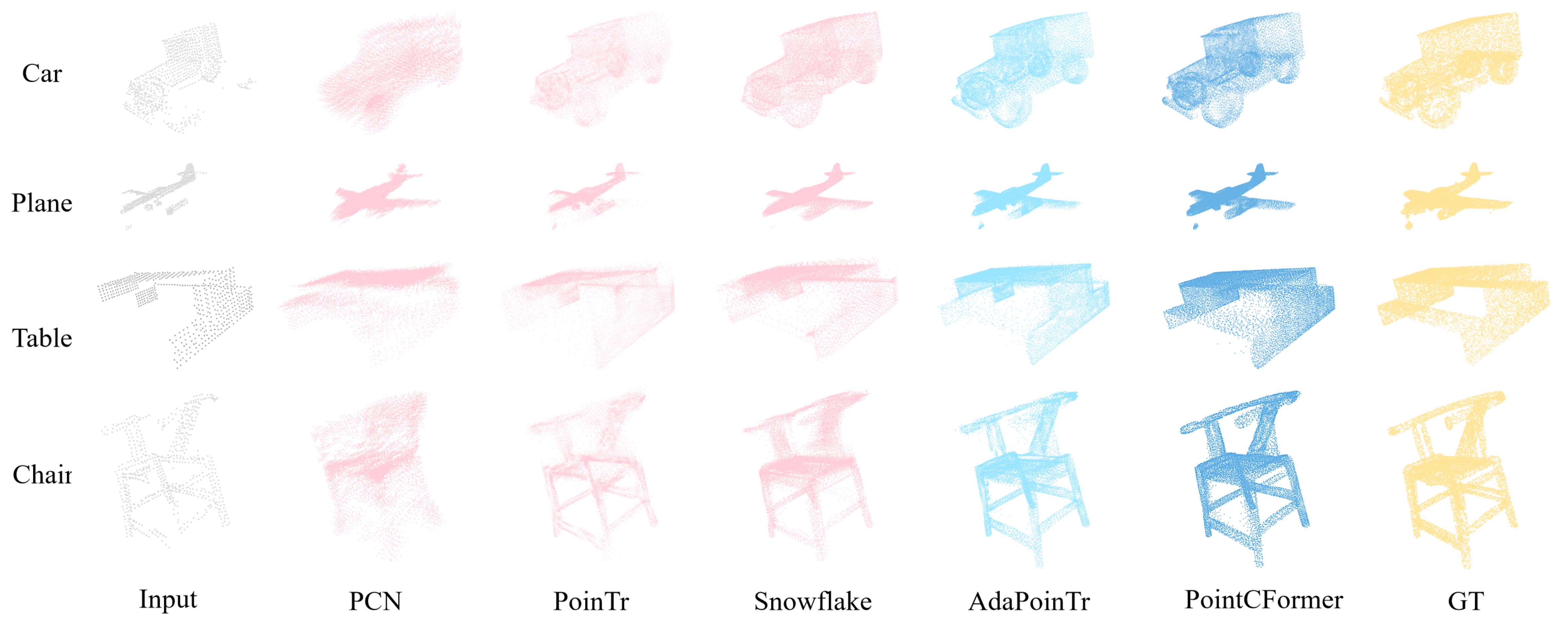}
\caption{Visual examples of point cloud completion results on the PCN dataset using different methods.}
\label{fig_pcn}
\end{figure*}

\subsubsection{Evaluation on PCN.}
The PCN dataset comprises 28,974 shapes for training and 1,200 shapes for testing, distributed across eight categories. It is currently the most frequently utilized benchmark dataset for evaluating the performance of point cloud completion methods. As shown in Table~\ref{table:pcn}, our method leads in almost all metrics. Specifically, our method improves the average CD-$\ell_1$ by 0.43 compared to the recently proposed PointAttN and shows remarkable improvements across different categories. Compared to the 3DMambaCom, our average CD-$\ell_1$ improves by 0.5, and we see a 0.31 increase in the F-Score@1\%.


Fig.~\ref{fig_pcn} illustrates visual comparisons of different methods on the PCN dataset. PointCFormer yields results that are closer to the ground truth. Specifically, the input point clouds shown in the top two rows exhibit unique features of their original shapes (i.e., cars and airplanes). While most adopted methods can reconstruct general 3D structures, our PointCFormer demonstrates advantages in capturing complex local details (e.g., the wings of an airplane and the wheels of a car). Moreover, when faced with the two sets of challenging point cloud inputs shown below, our PointCFormer achieves the most accurate recovery of the overall shape, outperforming other methods in both fidelity and accuracy.


\begin{table}[t]
\caption{Point cloud completion results on ShapeNet-55.}
\centering
\resizebox{\columnwidth}{!}{
    \begin{tabular}{c|ccc|cc}
        \toprule[0.5mm]
        Method & CD-$\ell_2$-S & CD-$\ell_2$-M & CD-$\ell_2$-H & CD-$\ell_2$-Avg & F-Score@1\%\\
        \midrule
        FoldingNet & 2.67 & 2.66 & 4.05 & 3.12 & 0.082\\
        PCN & 1.94 & 1.96 & 4.08 & 2.66 & 0.133\\
        PoinTr & 0.67 & 1.05 & 2.02 & 1.25 & 0.446\\
        Snowflake & 0.81 & 1.17 & 2.20 & 1.40 & 0.343\\
        AnchorFormer & 1.14 & 1.12 & 1.91 & 1.39 & 0.327\\
        PointAttN & 0.47 & 0.66 & 1.17 & 0.77 & -\\
        3DMambCom & 0.61 & 0.77 & 1.20 & 0.86 & 0.341\\
        AdaPoinTr & 0.49 & 0.69 & 1.24 & 0.81 & \textbf{0.503}\\
        PointCFormer & \textbf{0.42} & \textbf{0.64} & \textbf{1.15} & \textbf{0.73} & 0.499\\
        \toprule[0.5mm]
\end{tabular}}
\label{table:shape55}
\end{table}

\subsubsection{Evaluation on ShapeNet-55.}
Next, we evaluate PointCFormer on the ShapeNet-55 dataset, which encompasses more categories. Table~\ref{table:shape55} presents the performance of various methods on incomplete point cloud data with three different missing ratios (CD-$\ell_2$-S, CD-$\ell_2$-M, and CD-$\ell_2$-H) in terms of L2 Chamfer distance (CD-$\ell_2$) and F-Score@1\%. PointCFormer evidently outperforms other existing methods in CD-$\ell_2$-S, CD-$\ell_2$-M, CD-$\ell_2$-H, and average CD-$\ell_2$. Despite scoring slightly lower than the top-performing method (AdaPoinTr) in terms of F-Score@1\%, we achieve state-of-the-art performance with a CD-$\ell_2$ of 0.73. These results underscore the effective completion capabilities of PointCFormer, even when applied to such a diverse dataset.

\begin{table*}[t]
\caption{Performance comparison on ShapeNet-34: results for three difficulty levels across 34 seen and 21 unseen categories. CD-S, CD-M, and CD-H represent CD-$\ell_2$ (multiplied by 1000) results under simple, medium, and hard settings, respectively.}
\label{table:shape34uq22}
\centering
\resizebox{0.9\linewidth}{!}{
\begin{tabular}{ccccccccccc}
\bottomrule[0.5mm]
 &\multicolumn{5}{c}{\textbf{\small34 seen categories}}&\multicolumn{5}{c}{\textbf{\small21 unseen categories}}\\
\cmidrule(lr){2-6} \cmidrule(lr){7-11}
 & CD-S & CD-M & CD-H & CD-$\ell_2$-Avg & F-Score@1\% & CD-S & CD-M & CD-H & CD-$\ell_2$-Avg & F-Score@1\%\\
\midrule
FoldingNet & 1.86&1.81&3.38&2.35&0.139&2.76 & 2.74 & 5.36 & 3.62 & 0.095\\
PCN & 1.87&1.81&2.97&2.22&0.154&3.17 & 3.08 & 5.29 & 3.85 & 0.101\\
PoinTr & 0.76 & 1.05 & 1.88 & 1.23 & 0.421 & 1.04 & 1.67 & 3.44 & 2.05 & 0.384\\
Snowflake & 0.60 & 0.86 & 1.50 & 0.99 & 0.422 &0.88 & 1.46 & 2.92 & 1.75 & 0.388\\
AnchorFormer & 0.85&1.09&1.77&1.23&0.328&1.09 & 1.58 & 2.88 & 1.85 & 0.289\\
PointAttN & 0.51&0.70&1.23&0.81&-&0.76 & 1.15 & 2.23 & 1.38 & -\\
3DMambCom & 0.66&0.84&1.39&0.96&0.324&0.86 & 1.34 & 2.99 & 1.73 & 0.281\\
AdaPoinTr & 0.48&0.63&1.07&0.73&\textbf{0.469}&0.61 & 0.96 & 2.11 & 1.23 & 0.416\\
\midrule
PointCFormer & \textbf{0.41}&\textbf{0.55}&\textbf{1.03}&\textbf{0.66}&0.459&\textbf{0.53} & \textbf{0.88} & \textbf{1.97} & \textbf{1.12} & \textbf{0.420}\\
\bottomrule[0.5mm]
\end{tabular}}
\end{table*}

\subsubsection{Evaluation on ShapeNet-34/Unseen-21}
We conduct experiments on ShapeNet-34/unseen21 to evaluate the generalization ability of PointCFormer. The results presented in Table~\ref{table:shape34uq22} indicate that despite a close F1 score with the runner-up under various ratio settings, PointCFormer consistently achieves the best performance on the CD metric across both the 34 visible categories and the 21 unseen categories. This reflects its excellent generalization ability. 

\begin{figure}[t]
\centering
\includegraphics[width=0.9\columnwidth]{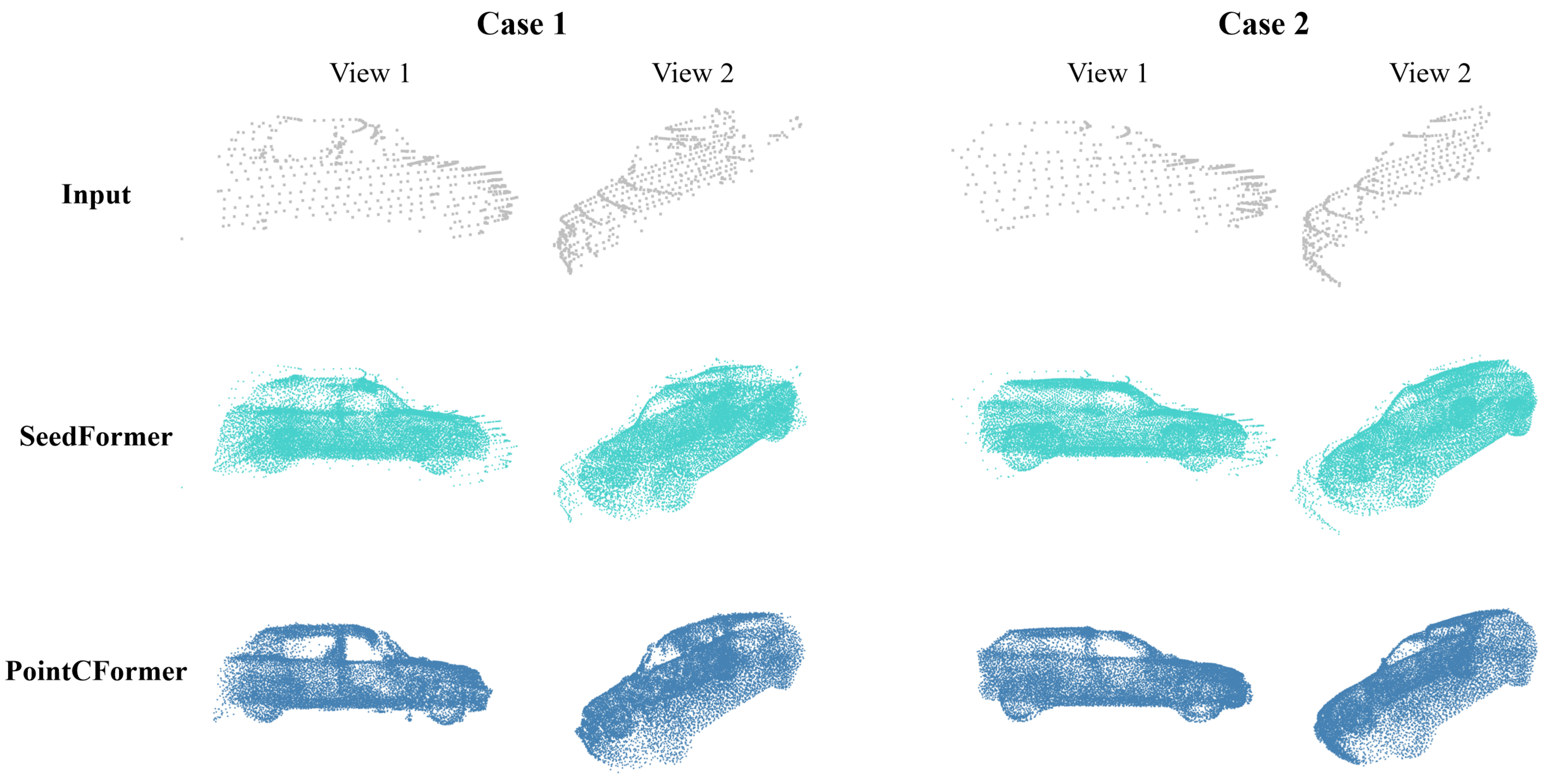}
\caption{Qualitative results on KITTI: dual-view representation of the same point cloud for enhanced visualization of car shapes in each scenario.}
\label{fig_kitti}
\end{figure}

\begin{table}[t]
\caption{Quantative comparison on KITTI in terms of MMD and Fidelity.}
\centering
\resizebox{\columnwidth}{!}{
    \begin{tabular}{c|cccc|c}
        \toprule[0.5mm]
        *1000 & SnowflakeNet & SeedFormer & PointAttN & AdaPoinTr & PointCFormer\\
        \midrule 
        Fidelity$\downarrow$ & 0.110 & 0.151 & 0.672 & 0.237 & \textbf{0.001}\\
        MMD$\downarrow$ & 0.907 & 0.516 & 0.504 & 0.392 & \textbf{0.353}\\
        \toprule[0.5mm]
\end{tabular}}
\label{table:kitti}
\end{table}

\subsubsection{Evaluation on KITTI}
KITTI dataset comprises incomplete point clouds of real vehicles captured by LiDAR scans of world scenes. Following the protocol in~\cite{xie2020grnet}, we fine-tune our model initially trained on ShapeNetCars~\cite{yuan2018pcn} and assess its performance on the KITTI dataset. As reported in Table~\ref{table:kitti}, PointCFormer consistently outperforms other models across the fidelity and MMD metrics. This illustrates the superior capability of PointCFormer in capturing the 3D shape characteristics of vehicles. Furthermore, Fig.~\ref{fig_kitti} shows two examples of point cloud completion across two views. PointCFormer exhibits superior overall quality with finer local details in granular patterns, reflecting the advantages of leveraging the novel feature extraction paradigm we have introduced for enhancing the network's ability to parse 3D structures.


\subsection{Ablation Study}

We conducted a comprehensive ablation study on the PCN dataset to demonstrate the effectiveness of each module we designed within PointCFormer. The results are summarized in Table~\ref{table_ab}. Here, ``LGRP'' refers to a progressive feature extractor that contains only two ST-LFE modules, ``GSP'' refers to a progressive feature extractor that uses only two self-attention layers (Applying LGRP and GSP is the whole PFE.), and ``CM'' stands for the correction module. The baseline ``I'' is a combination of the partial framework from AdaPoinTr with a conventional encoder. We observe a significant performance boost when we replace the traditional local feature extraction process with ``LGRP''. The model’s performance was further enhanced by incorporating the global shape perception process (GSP). Lastly, the model's best performance was obtained after adding the correction module. It is evident from the results that the local geometric relationship perception contributes the most to the model's improvement, showcasing its great potential for other point cloud tasks.

\begin{table}[t]
    \caption{Ablation study of several modules proposed in PointCFormer on the PCN Dataset.}
  \centering
  \small
  \resizebox{0.95\columnwidth}{!}{
  \begin{tabular}{c | c c c | c c}
    \toprule
    & LGRP & GSP & CM & CD-$\ell_1$$\downarrow$ & F-Score@1\%$ \uparrow$\\
    \midrule
    \textrm{I}   & &  & & 6.92 & 0.810	\\
    \textrm{II}  & \checkmark & &  & 6.47 & 0.840\\
    \textrm{III} & & \checkmark &  & 6.85 & 0.822\\
    \textrm{IV} & & & \checkmark & 6.87 & 0.819\\
    \midrule
    \textrm{V}  & & \checkmark & \checkmark & 6.81 & 0.828\\
    \textrm{VI}&\checkmark & \checkmark &  & 6.44 & 0.853\\
    \textrm{VII}&\checkmark &  & \checkmark & 6.48 & 0.844\\
    \textrm{VIII}&\checkmark & \checkmark & \checkmark & \textbf{6.41} & \textbf{0.855}\\
    \bottomrule
    \end{tabular}}
    \label{table_ab}
\end{table}

\section{Conclusions}
In this paper, we explore preserving extensive global structural features while extracting detailed local features in point cloud completion tasks. We introduce PointCFormer, a novel point cloud completion model that builds upon the Transformer framework. A key component is the local relation perception module, which measures the contribution of each point within a local region based on the relations in 3D space and high-dimensional feature space. Another key component is the progressive feature extractor combined with the attention mechanism to strengthen the capture of global structures. Additionally, the latent space correction module ensures that the generated point proxies align more accurately with the distribution of the original input point cloud data. Crucially, we incorporate a relation-based local perception mechanism into both the feature extractor and the correction module, significantly enhancing their performance. Comprehensive comparative and ablation studies across various challenging benchmarks demonstrate that PointCFormer achieves superior completion performance. In fact, our relation-based feature extraction method comprehensively considers the global and local information of point clouds. We would like to extend our method to other point cloud-related tasks in the future.

\section{Acknowledgments}
This work is partially funded by the National Natural Science Foundation of China (12494550 and 12494553), the Beijing Natural Science Foundation (Z240002), and  the Guangdong Science and Technology Program (2023B1515120026).





\bibliography{aaai25}

\end{document}


\maketitle

In this supplementary material, we provide:
\begin{itemize}
\item Detailed experimental procedures and hyper-parameter recommendations for training PointCFormer.
\item Extensive explanations and comparative analysis of the differences between R1 and R2 relation measures under comprehensive ablation studies.
\item More visual and numerical results of PointCFormer on various baselines.
\item Complexity analysis of PointCFormer.
\end{itemize}

\section{Hyper-parameter Recommendation}

\subsection{Base setting}
Similar to the AdapoinTr~\cite{10232862} protocol standard, we have set the depth of the encoders and decoders in the Transformer to 6 and 8, respectively. We have employed 6-head attention across all Transformer blocks and have set the hidden dimension to 384.

\subsection{Progressive Feature Extractor}
We utilize a lightweight DGCNN~\cite{wang2019dynamic} feature extractor with a single-layer structure for initial coarse extraction, with the k of the geometric perception kNN operation set to 16.

In experiments conducted on the PCN~\cite{yuan2018pcn} and KITTI datasets, the network takes 2048 points as input and completes them to 16384 points. We have set the batch size to 16 and trained the model for 600 epochs, with a continuous learning rate decay of 0.9 every 20 epochs. In the first stage of the feature extractor, the DGCNN combined with FPS on such datasets initially generates 1048 dense point proxies and their 96-dimensional features from the original input. In the second stage, the two instances of ST-LFE respectively reduce the scale of processed points to 512 and 256, and upscale the features to 192 and 384 dimensions.

On the ShapeNet-55/34~\cite{yu2021pointr} and related datasets, the model takes 2048 points as input and completes them to 8192 points. We have set the batch size to 48. In the first stage of the feature extractor, the DGCNN combined with FPS on such datasets initially generates 640 dense point proxies and their features from the original input. In the second stage, the ST-LFE respectively reduces the scale of processed points to 256 and 128, and upscale the features to 192 and 384 dimensions.

\subsection{Generated Point Proxy Correction Module}
On the PCN and ShapeNet-55 datasets, we reintroduce 1536 and 768 high-density input points respectively, which are then connected with the generated point proxies to form a dense input for the correction module. This is then reduced back to the original scale of the generated point proxies through two instances of ST-LFE (on the PCN dataset, ST-LFE reduces the number of processed points by 1024 and 512 in two separate instances. Similarly, on the ShapeNet-55 dataset, ST-LFE reduces the number of processed points by 512 and 256 in two separate instances). During this process, the feature dimensions processed by ST-LFE and the self-attention head are kept below 32, ensuring computational efficiency through low-dimensional processing.
\begin{table}[t]
\caption{Comparative experiment of different local perception methods with various k sizes in kNN on PCN.}
\centering
\resizebox{\columnwidth}{!}{
    \begin{tabular}{cc|ccc|cc}
        \toprule[0.5mm]
        Perceptor & k Size & Plane & Car & Lamp & Avg CD-$\ell_1\downarrow$ & F-Score@1\%$\uparrow$\\
        \hline
        \multirow{3}{*}{EdgeConv} & 8 & 3.68 & 7.47 & 5.47 & 6.52 & 0.846\\
        & 16 & 3.65 & 7.37 & 5.47 & 6.50 & 0.847\\
        & 24 & 3.75 & 7.46 & 5.50 & 6.54 & 0.846\\
        \midrule
        \multirow{3}{*}{LGRP} & 8 & 3.55 & 7.37 & 5.16 & 6.46 & 0.851\\
        & 16 & 3.53 & 7.32 & 5.12 & 6.41 & 0.855\\
        & 24 & 3.49 & 7.34 & 5.10 & 6.39 & 0.857\\
        \toprule[0.5mm]
\end{tabular}}
\label{table:ksize}
\end{table}

\subsection{Local Geometric Relationship Perception}
In our comparative experimental analysis on the PCN dataset, we focus on selecting \( k \) in the kNN method within Local Geometric Relationship Perception (LGRP). The results, as shown in Table~\ref{table:ksize}, demonstrated that using EdgeConv as the local preceptor for feature extraction throughout the network does not necessarily lead to improved performance with an increase in \( k \). In fact, beyond a certain threshold, the performance of the network began to deteriorate slightly. This supports our hypothesis that ``in the context of sparse point inputs, the local perception process may be negatively impacted by an excessive focus on less relevant or negatively correlated neighboring points, which could introduce errors into the feature extraction process.'' To mitigate this adverse effect, we proposed LGRP. By incorporating a consideration of relational degrees, LGRP can autonomously filter out irrelevant points, thereby enhancing the feature extraction process. The results suggest that, within a substantial range, an increase in the value of \( k \) does not negatively impact the network's performance. Therefore, balancing both performance and computational efficiency, we have chosen 16 as the optimal value for \( k \) in the kNN perception within the LGRP.

\begin{figure}[t]
\centering
\includegraphics[width=1\columnwidth]{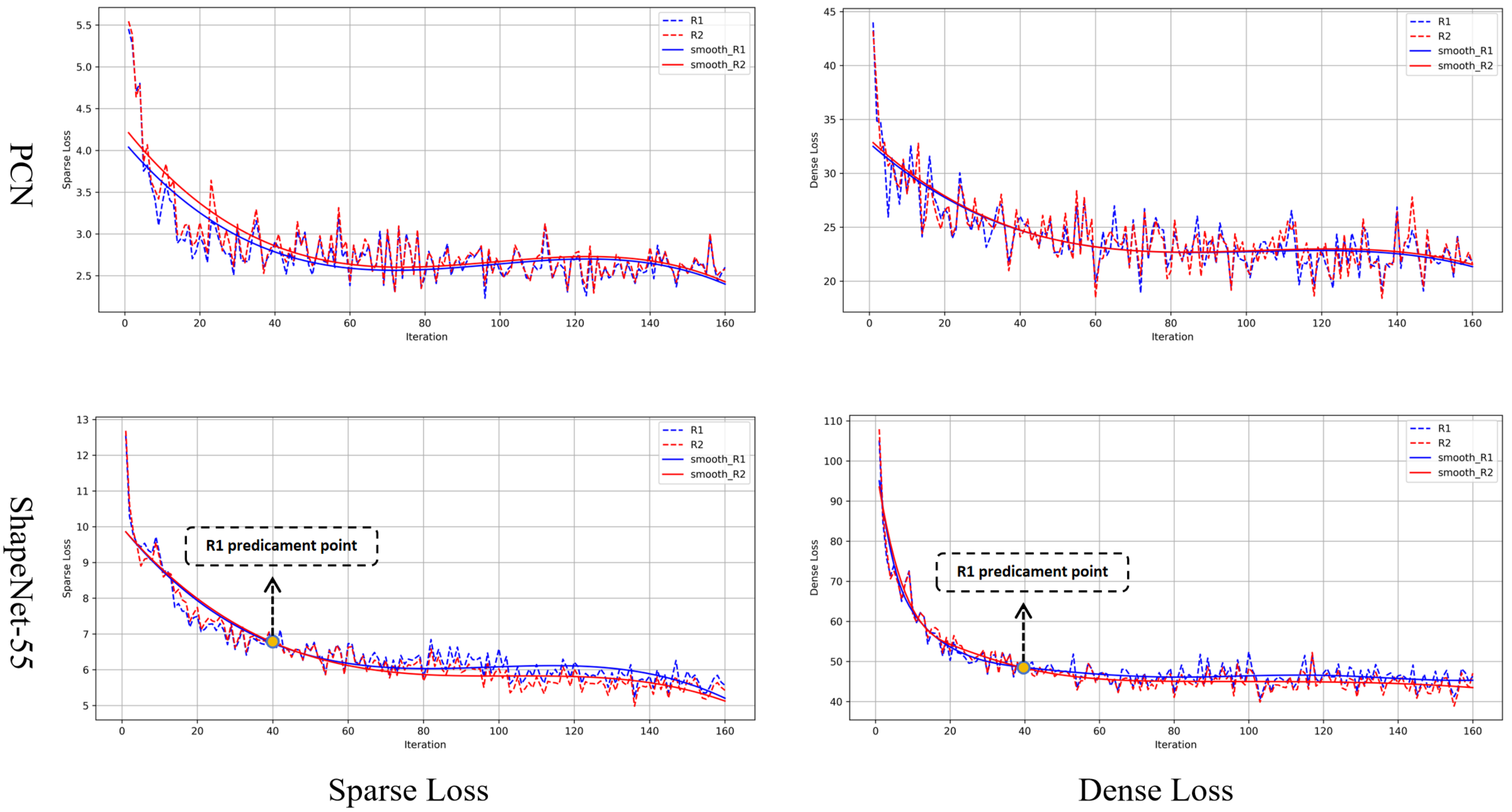}
\caption{Training loss curves using individual R1 and R2 relational metrics on PCN and ShapeNet-55. ``Dense Loss'' reflects overall shape completion quality, and ``Sparse Loss'' indicates the quality of generated point proxies.}
\label{fig_los}
\end{figure}

\section{Full Analysis of R1 and R2}

\subsection{Applicable Scenarios}
To further analyze the applicability of R1 and R2 relational metrics and their differential impact on local feature perception, we conducted experiments on the PCN and ShapeNet-55 datasets, with the respective loss curves displayed in Fig.~\ref{fig_los}. The first row of the figure presents the training curves on the PCN dataset, where we observe that the Sparse Loss decreases rapidly when using the R1 relational metric and maintains a slightly lower loss compared to the R2 curve throughout the subsequent training. In contrast, the Dense Loss trends for both R1 and R2 are nearly identical. This phenomenon occurs because R1 represents a rigid three-dimensional spatial constraint that is independent of the neural network's feature relationships, allowing point proxies to quickly identify spatial geometric relationships and converge swiftly.

However, the second row of the figure, which shows the training curves on the ShapeNet-55 dataset, demonstrates the superiority of R2. R2 is adaptable to nearly any data distribution due to its dynamic nature, stemming from the inter-feature relationships within the network. The variance in point cloud distributions means that strong, rigid constraints like those imposed by R1 may not be suitable, as evidenced by the R1 curve reaching a plateau and seemingly entering a training impasse. This plateau is attributed to the strong constraints introduced by R1, whereas R2 does not encounter this issue.

An analysis of these observations suggests that while R1 rapidly establishes geometric relationships in data with a consistent spatial structure, it may falter when faced with diverse distributions that do not adhere to its rigid constraints. On the other hand, R2's flexible, feature-based relational approach allows it to adapt and excel across varying point cloud distributions, making it a more universally applicable metric in the context of point cloud completion. In this paper, we use the R1 and R2 relational metrics simultaneously.

\begin{table}[t]
\caption{Performance comparison of pointCFormer using varied combinations of R1 and R2 relational metrics across multiple datasets.}
\centering
\resizebox{\columnwidth}{!}{
    \begin{tabular}{cc|ccc}
        \toprule[0.5mm]
        Dataset & Setting & Avg CD-$\ell_1\downarrow$ & Avg CD-$\ell_2\downarrow$ & F-Score@1\%$\uparrow$\\
        \hline
        \multirow{4}{*}{PCN} & Base & 6.52 & - & 0.846\\
        & R1 & 6.44 & - & 0.853\\
        & R2 & 6.43 & - & 0.852\\
        & R1\&R2 & 6.41 & - & 0.855\\
        \midrule
        \multirow{4}{*}{ShapeNet-55} & Base & - & 0.83 & 0.485\\
        & R1 & - & 0.89 & 0.478\\
        & R2 & - & 0.75 & 0.501\\
        & R1\&R2 & - & 0.73 & 0.499\\
        \midrule
        \multirow{4}{*}{ShapeNet-34} & Base & - & 0.84 & 0.441\\
        & R1 & - & 0.70 & 0.455\\
        & R2 & - & 0.69 & 0.458\\
        & R1\&R2 & - & 0.66 & 0.459\\
        \toprule[0.5mm]
\end{tabular}}
\label{table:r1r2c}
\end{table}

\subsection{Performance Comparison}
Quantitative experimental results, as shown in Table~\ref{table:r1r2c}, indicate that on the PCN dataset, there is no significant difference between the performance of R1 and R2, both offering substantial improvements over the baseline. However, on the ShapeNet-55/34 datasets, employing R1 alone introduces strong spatial constraints that can result in performance worse than the baseline, while using R2 alone does not have this adverse effect. The combination of R1 and R2 largely mitigates this issue and ensures continued optimization and enhancement of the model.

\section{Detailed Experimental Results}

\subsection{Visualization results on PCN}
\begin{figure*}[t]
\centering
\includegraphics[width=1\textwidth]{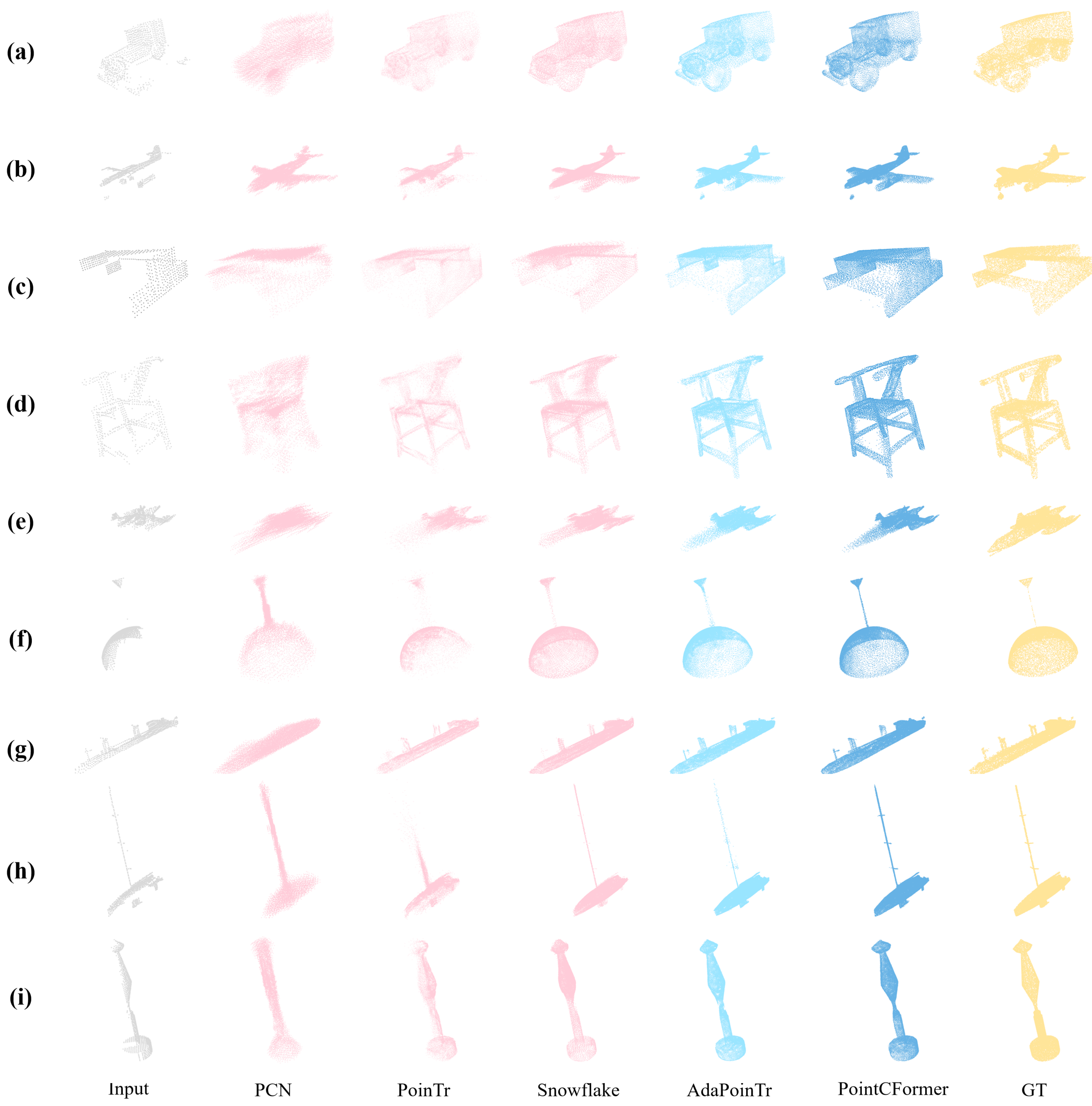}
\caption{More qualitative results on PCN.}
\label{fig_suppcn}
\end{figure*}

Fig.~\ref{fig_suppcn} presents additional qualitative results from the PCN dataset. Our results visually outperform the baseline methods in both overall shape and local detail. As illustrated in Table~\ref{table:com_adapointr}, our methodology, while drawing elements from AdaPoinTr, is not simply an incremental model. Instead, it shows remarkable improvement in overall performance. This is largely due to our innovative relation-based local feature perception and the progressive optimization in feature extraction.

\begin{table}[t]
\caption{Further comparison of our method with AdaPoinTr and its expanded-size version.}
\centering
\resizebox{\columnwidth}{!}{
    \begin{tabular}{c|cc|cc}
        \toprule[0.5mm]
        Method & Params & FLOPs & Avg CD-$\ell_1$ & F-Score@1\%\\
        \midrule
        AdaPoinTr & 32.4M & 241G & 6.54 & 0.841\\
        AdaPoinTr-Large & 39.6M & 271G & 6.49 & 0.847\\
        PointCFormer & 33.2M & 274G & \textbf{6.41} & \textbf{0.855}\\
        \toprule[0.5mm]
\end{tabular}}
\label{table:com_adapointr}
\end{table}

\subsection{Visualization results on ShapeNet-55}
\begin{figure*}[t]
\centering
\includegraphics[width=1\textwidth]{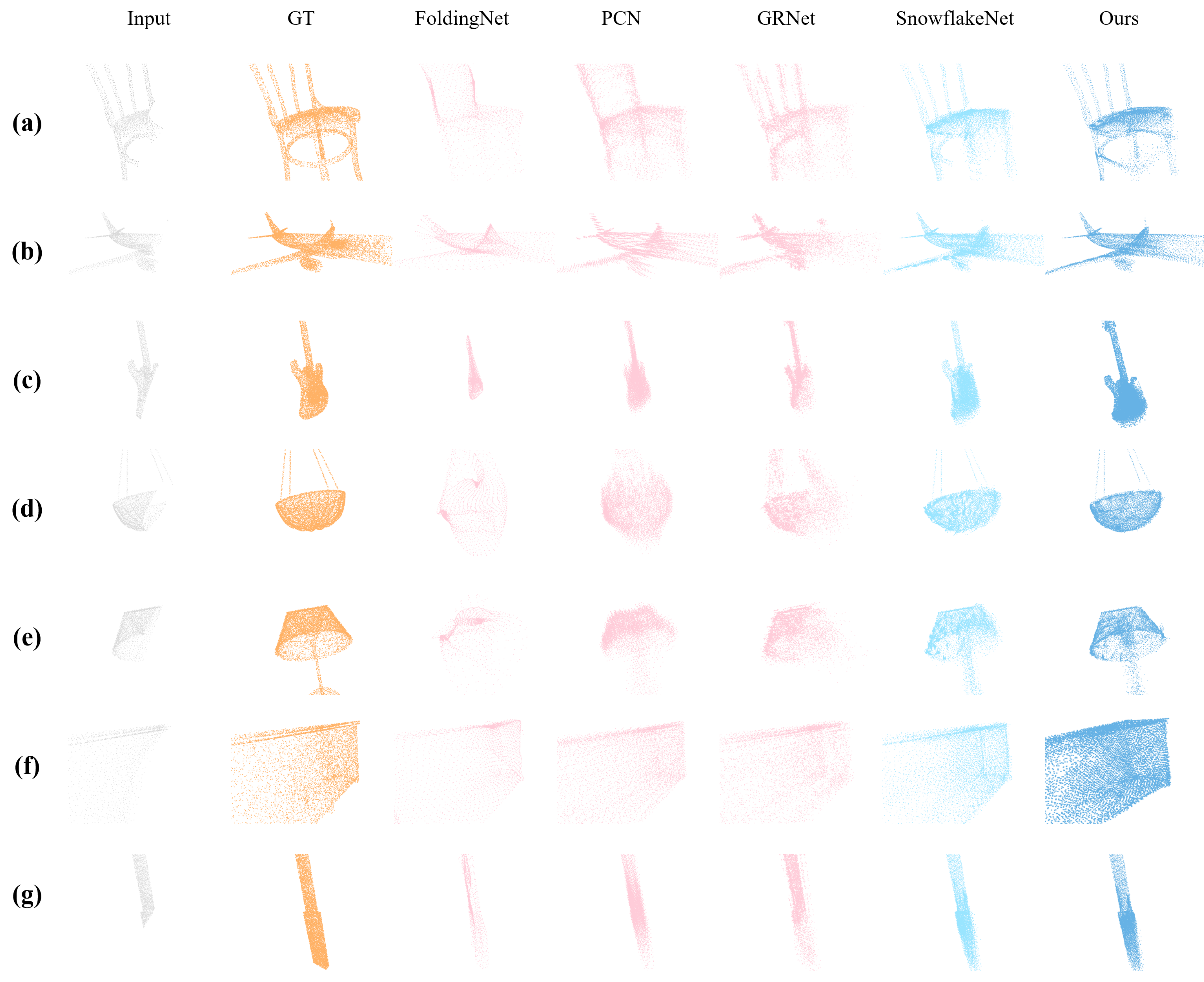}
\caption{More qualitative results on ShapeNet-55.}
\label{fig_supshape55}
\end{figure*}

Fig.~\ref{fig_supshape55} showcases the visualization results from the ShapeNet-55 dataset. Our results exhibit a marked visual improvement over the baseline method.

\subsection{Visualization results on KITTI}
\begin{figure*}[t]
\centering
\includegraphics[width=1\textwidth]{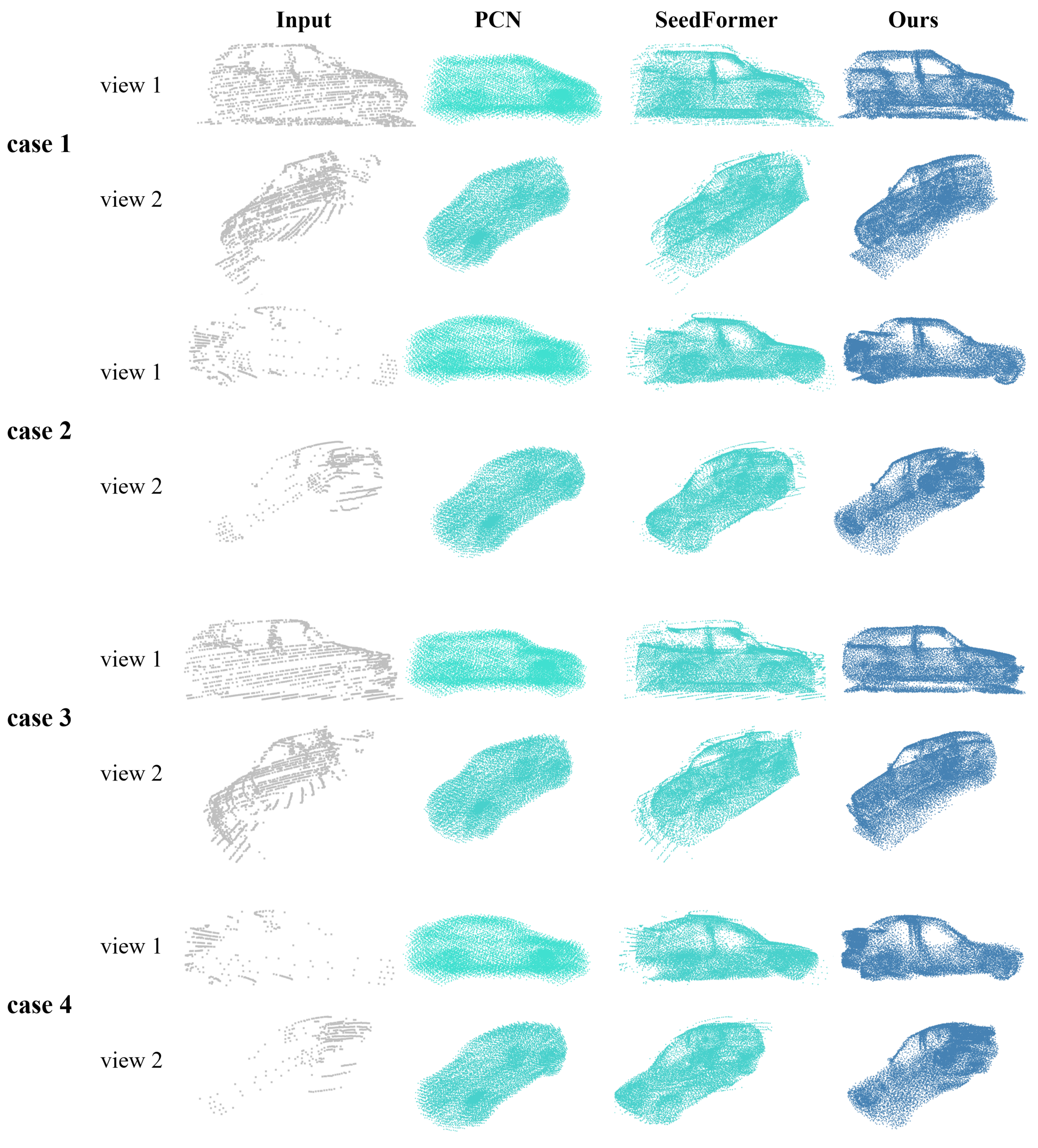}
\caption{More qualitative results on KITTI.}
\label{fig_supkitti}
\end{figure*}

Fig.~\ref{fig_supkitti} presents the visualization results derived from the KITTI dataset. In multiple examples, our results demonstrate a closer alignment with real-world visual features from various viewpoints.

\begin{table*}[t]
\caption{Results on Projected-ShapeNet-55: we report the detailed results for each method on 10 categories and the overall results on 55 categories under CD-$\ell_1$(multiplied by 1000). We also report F-Score@1\% metric.}
\label{table:proshape55}
\centering
\resizebox{\linewidth}{!}{
    \begin{tabular}{c|ccccc|ccccc|cc}
        \toprule[0.5mm]
        & Table & Chair & Air & Car & Sofa & Bird house & Bag & Remote & Key board & Rocket & CD-$\ell_1$-Avg$\downarrow$ & F-Score@1\%$\uparrow$\\
        \midrule
        PCN   & 14.79 & 15.33 & 9.07 & 12.85 & 17.12 & 20.38 & 18.64 & 14.62 & 13.69 & 10.98 & 16.64 & 0.403 \\
        TopNet & 14.40 & 16.29 & 9.85 & 13.61 & 16.93 & 22.00 & 18.69 & 13.52 & 11.05 & 10.45 & 16.35 & 0.337 \\
        GRNet & 12.01 & 12.57 & 8.30 & 12.13 & 14.36 & 16.52 & 14.67 & 12.18 & 9.71  & 8.58  & 12.81 & 0.491 \\
        SnowflakeNet & 10.49 & 11.07 & 6.35 & 11.20 & 12.59 & 15.24 & 12.86 & 10.07 & 8.12  & 7.49  & 11.34 & 0.594 \\
        PoinTr & 9.97  & 10.43 & 6.02 & 10.58 & 12.11 & 14.60 & 12.15 & 9.55  & 7.61  & 6.86  & 10.68 & 0.615 \\
        AdaPoinTr & 8.81  & 9.12  & 5.18 & 9.77  & 10.89 & 13.27 & \textbf{10.93} & 8.81  & 6.79  & 5.58  & 9.58  & 0.701 \\
        \midrule
        PointCFormer & \textbf{8.55}  & \textbf{8.90}  & \textbf{5.01} & \textbf{9.57}  & \textbf{10.65} & \textbf{12.96} & 10.99 & \textbf{8.42}  & \textbf{6.61}  & \textbf{5.42}  & \textbf{9.29}  & \textbf{0.713} \\
        PointCFormer-Plus & -  & -  & - & -  & - & - & - & -  & -  & -  & \textbf{9.07}  & \textbf{0.728} \\
        \toprule[0.5mm]
\end{tabular}}
\end{table*}

\begin{table*}[t]
\caption{Performance comparison on Projected-ShapeNet-34: we report the results under CD-$\ell_1$(multiplied by 1000) of 34 seen categories and 21 unseen categories, and we also report F-Score@1\% metric for each method.}
\label{table:proshape34}
\centering
\resizebox{\linewidth}{!}{
\begin{tabular}{cccc|ccccc|cc}
\bottomrule[0.5mm]
 &\multicolumn{5}{c}{\textbf{\small34 seen categories}}&\multicolumn{5}{c}{\textbf{\small21 unseen categories}}\\
\cmidrule(lr){2-6} \cmidrule(lr){7-11}
 & Bin & Knife & Table & CD-$\ell_1$$\downarrow$ & F-Score@1\%$\uparrow$ & Microphone & Skateboard & Earphone & CD-$\ell_1$$\downarrow$ & F-Score@1\%$\uparrow$\\
\midrule
PCN   & 17.60 & 8.52 & 13.69 & 15.53 & 0.432 & 18.05 & 17.27 & 24.82 & 21.44 & 0.307 \\
TopNet & 14.90 & 7.58 & 11.18 & 12.96 & 0.464 & 14.34 & 12.59 & 19.34 & 15.98 & 0.358 \\
GRNet & 14.79 & 7.84 & 11.00 & 12.41 & 0.506 & 11.39 & 10.60 & 15.00 & 15.03 & 0.439 \\
SnowflakeNet & 13.21 & 5.80 & 9.46 & 10.69 & 0.616 & 10.10 & 9.58 & 15.19 & 12.82 & 0.551 \\
PoinTr & 12.36 & 5.64 & 8.97 & 10.21 & 0.634 & 9.34 & 8.98 & 14.23 & 12.43 & 0.555 \\
AdaPoinTr & 11.45 & 4.95 & 7.95 & 9.12 & 0.721 & 7.96 & 8.34 & 12.30 & 11.37 & 0.642 \\
\midrule
PointCFormer & \textbf{11.09} & \textbf{4.66} & \textbf{7.81} & \textbf{8.90} & \textbf{0.738} & \textbf{7.75} & \textbf{8.09} & \textbf{12.08} & \textbf{11.04} & \textbf{0.657} \\
PointCFormer-Plus & - & - & - & - & - & - & - & - & \textbf{10.71} & \textbf{0.680} \\
\bottomrule[0.5mm]
\end{tabular}}
\end{table*}

\subsection{Detailed Results on Additional Datasets}
\subsubsection{Projected-ShapeNet-55}
We conduct experiments on Projected-ShapeNet-55~\cite{10232862}, a dataset where the input point clouds are generated through a noise back-projection method, against several state-of-the-art approaches~\cite{yuan2018pcn,tchapmi2019topnet,xie2020grnet,xiang2021snowflakenet,yu2021pointr,10232862}. The input point clouds in Projected-ShapeNet-55 contain noise, which makes the dataset more ambiguous, challenging, and realistic. We report the results of our model in comparison to other existing methods. We document the class-wise Chamfer Distance (CD) and the overall CD for all methods. To illustrate detailed results, we select ten representative categories. As shown in Table~\ref{table:proshape55}, our PointCFormer achieves the best performance across the ten categories and in overall CD, with an improvement of 0.29 in CD-$\ell_1$ (multiplied by 1000) and 0.012 in F-Score@1\% compared to AdaPoinTr.

\subsubsection{Projected-ShapeNet-34}
We also validate the performance of PointCFormer on Project-ShapeNet-34, as shown in Table~\ref{table:proshape34}. On Project-ShapeNet-34, the model is trained on the segmentation of 34 seen categories and then tested on these 34 seen categories as well as 21 unseen categories during testing. PointCFormer achieves state-of-the-art performance on both the 34 seen and 21 unseen categories.

\section{Complexity Analysis}
Table~\ref{table:comp} provides a detailed complexity analysis of our method. We report the theoretical computational costs (FLOPs) and throughput (point clouds per second) of our method compared to other approaches. We also present the results on the ShapeNet-55, the unseen categories within ShapeNet34, and the PCN benchmark using the average Chamfer distance as a metric for evaluation. Compared with AdaPoinTr, our method only exhibits a minor increase in FLOPs while maintaining stable throughput. This demonstrates that our method strikes an effective balance between performance and efficiency.

\begin{table}[t]
\caption{Complexity analysis. We report the theoretical computation cost (FLOPs) and throughput (T.put) of our method and existing methods.}
\centering
\resizebox{\columnwidth}{!}{
    \begin{tabular}{c|cc|cc|c}
        \toprule[0.5mm]
        Method & FLOPs & T.put & CD$_{55}$ & CD$_{34}$ & CD$_{PCN}$\\
        \midrule 
        FoldingNet & 27.58G & 158 pc/s & 3.12 & 3.62 & 14.31\\
        GRNet & 40.44G & 65 pc/s & 1.97 & 2.99 & 8.83\\
        SnowflakeNet & 17.16G & 72 pc/s & 1.40 & 1.75 & 7.21\\
        SeedFormer & 13.97G & 21 pc/s & 0.92 & 1.34 & 6.74\\
        AdaPoinTr & 12.43G & 61 pc/s & 0.81 & 1.23 & 6.53\\
        PointCFormer & 15.81G & 52 pc/s & \textbf{0.73} & \textbf{1.12} & \textbf{6.41}\\
        \toprule[0.5mm]
\end{tabular}}
\label{table:comp}
\end{table}

\bibliography{aaai25}